\documentclass{Interspeech2024}

\interspeechcameraready

\title{CNVSRC 2023: The First Chinese Continuous Visual Speech Recognition Challenge}
\name[affiliation={1}]{Chen}{Chen}
\name[affiliation={2}]{Zehua}{Liu}
\name[affiliation={2}]{Xiaolou}{Li}
\name[affiliation={2}]{Lantian}{Li}
\name[affiliation={1}]{Dong}{Wang}
\address{
  $^1$Center for Speech and Language Technologies, BNRist, Tsinghua University, China\\
  $^2$Beijing University of Posts and Telecommunications, China
  \thanks{This work was supported by the National Natural Science Foundation of China (NSFC) under Grants No.62301075/62171250. L. Li and D. Wang are the corresponding authors.}
  }
\email{\{chenc21,wangdong99\}@mails.tsinghua.edu.cn, \{lixiaolou,liuzehua,lilt\}@bupt.edu.cn}


\begin{document}

\maketitle

\begin{abstract}

The first Chinese Continuous Visual Speech Recognition Challenge aimed to probe the performance of Large Vocabulary Continuous Visual Speech Recognition (LVC-VSR) on two tasks: (1) Single-speaker VSR for a particular speaker and (2) Multi-speaker VSR for a set of registered speakers.
The challenge yielded highly successful results, with the best submission significantly outperforming the baseline, particularly in the single-speaker task.
This paper comprehensively reviews the challenge, encompassing the data profile, task specifications, and baseline system construction.
It also summarises the representative techniques employed by the submitted systems, highlighting the most effective approaches. Additional information and resources about this challenge can be accessed through the official website at \emph{\url{http://cnceleb.org/competition}}.

\end{abstract}

\section{Introduction}

Visual Speech Recognition (VSR), commonly called lip reading, is a technology that utilizes lip movements to infer speech content.
It has broad applications, including public surveillance, support for elderly and disabled individuals, and fake video detection.
Traditionally, VSR has been primarily focused on recognizing isolated words or phrases.
For example, Martinez et al.~\cite{martinez_lipreading_2020} 
developed a model that extracts visual features using 3D convolution, ResNet-18, and a multi-scale temporal convolutional network (MS-TCN). This was further enhanced by simple average pooling and a softmax layer for inferring word posteriors, resulting in commendable performance on the LRW~\cite{chung2017lip} and LRW-1000 datasets~\cite{yang2019lrw}, 
which are the largest publicly available benchmark datasets for unconstrained isolated word lip-reading in English and Mandarin, respectively.
Ma et al.~\cite{ma_lip-reading_2021} adopted a similar architecture but introduced a densely connected temporal convolutional network (DC-TCN) to achieve improved performance.


Recently, research in lip reading has progressed beyond word and phrase recognition to focus on large vocabulary continuous visual speech recognition (LVC-VSR),
a more challenging task but with more realistic merit. 
Significant advancements have been made in English benchmarks, partly attributable to the availability of large-scale English visual-speech datasets such as LRS2~\cite{afouras2018deep}, LRS3~\cite{afouras2018deep}, AVSpeech~\cite{ephrat_looking_2018}, and VoxCeleb~\cite{nagrani_voxceleb_2017,chung_voxceleb2_2018}.
While earlier studies addressed this issue using hand-crafted features and sequential models like HMM~\cite{goldschen1997continuous} or RNN~\cite{petridis2016deep}, 
a significant breakthrough occurred with the end-to-end approach, which processes raw video frames and generates a word sequence. LipNet~\cite{assael2016lipnet} is perhaps the first end-to-end model, integrating spatiotemporal convolution layers and bi-directional gated recurrent unit (BGRU), trained using the connectionist temporal classification(CTC) loss~\cite{graves_connectionist_2006}. 
The model underwent testing on GRID, a dataset with limited grammar and vocabulary ~\cite{cooke_audio-visual_2006}. A similar architecture was adopted by Jeon et al.~\cite{jeon_lipreading_2022}, with their approach involving the integration of multiple CNN-based streams into the feature extraction process.

While promising, the GRID corpus is limited in its ability to capture real-world complexity due to the constrained grammar and vocabulary of the sentences.
A more challenging task involves large vocabulary continuous visual speech recognition based on datasets gathered from online media repositories like YouTube. A seminal work~\cite{afouras2018deep}introduced the first comprehensive visual speech datasets LRS2 and LRS3, along with the first transformer-based system trained with either the CTC loss (TM-CTC) or the sequence-to-sequence loss (TM-seq2seq). 
In a parallel endeavour, Google~\cite{shillingford2018large} developed an end-to-end model (CNN/BLSTM backbone and CTC loss) that transcribes videos into phone sequences and utilizes an FST-based decoder to obtain word sequences.
The research was expanded upon in a subsequent study~\cite{ma_end--end_2021}, which introduced a hybrid CTC/Attention model utilizing a ResNet/Conformer encoder and a Transformer-based language model (LM). This work was further developed by incorporating time-masking data augmentation and an auxiliary reconstruction loss in a subsequent publication~\cite{ma_visual_2022}.

In addition to employing complex structures, a simple yet effective approach is increasing the training data volume. However, a major challenge arises from the dearth of well-labeled data. One promising strategy to address this issue involves utilizing a pre-trained automatic speech recognition (ASR) model to transcribe unlabelled videos. This methodology was extensively explored in~\cite{ma_auto-avsr_2023}, wherein a ResNet/Conformer model was trained on both unlabelled datasets like  
VoxCeleb2~\cite{chung_voxceleb2_2018} and AVSpeech~\cite{ephrat_looking_2018},
as well as text-labelled datasets such as LRW~\cite{chung2017lip}, LRS2~\cite{afouras2018deep}, and LRS3~\cite{afouras2018deep}. 
The study revealed significant performance enhancements with auto-labelled data, and further improvements were observed with increased incorporation of unlabelled data.

For VSR on Chinese data, the research progress has been significantly impeded by the scarcity of data resources. 
Despite the presence of LRW-1000~\cite{yang2019lrw} as the sole large-scale dataset in Chinese, it consists solely of isolated words.
In 2023, the release of the CN-CVS dataset~\cite{chen_cn-cvs_2023} marked the debut of the first large-scale continuous visual-speech dataset in Chinese, thereby presenting an opportunity to propel research in Chinese LVC-VSR.
This also motivated the first Chinese Continuous Visual Speech Recognition Challenge (CNVSRC 2023), hosted as a special session on NCMMSC 2023 to estimate the performance boundary of existing LVC-VSR techniques with Chinese data
and attracting research interest in this domain.
The CN-CVS~\cite{chen_cn-cvs_2023} dataset was used as the primary training data, supplemented by two additional datasets - CNVSRC-Single and CNVSRC-Multi - introduced by the organizers to facilitate system development and evaluation.
The organizers also published the model and code of the baseline systems so that the participants could use them as references when developing their systems. 
This paper summarises the challenge, emphasizing the predominant findings gleaned from the submitted systems.

The structure of the rest of this paper is outlined as follows:
Section~\ref{sec:task_and_data} introduces the tasks and data of the challenge.
In Section~\ref{sec:baseline}, a comprehensive description of the baseline system is provided, covering the model structure, training strategies, and performance evaluation.
Section~\ref{sec:result} reports the challenge result and summarizes the representative technologies utilized by the participants. 
Lastly, the paper is concluded in Section~\ref{sec:conclusion}.

\vspace{-1mm}
\section{Tasks and Data}
\label{sec:task_and_data}

\subsection{Tasks}
\vspace{-1mm}

The CNVSRC 2023 challenge encompasses two distinct tasks: Single-speaker VSR (T1) and Multi-speaker VSR (T2).
Task T1 emphasizes the performance of large-scale tuning for a specific speaker,
whereas T2 focuses on the fundamental performance of the system for non-specific but \emph{registered} speakers, 
i.e., speakers seen in the data for system development.
In both tasks, the system is fed with silent facial videos featuring a single individual, and it is required to generate the spoken content in written form.

Each task is further categorized into a `fixed track' and an `open track'. The fixed track permits the use of data and additional resources that have been agreed upon by the organizing committee. 
Conversely, the open track allows participants to employ any resources except the evaluation set.

Character Error Rate (CER) was used as the main metric to evaluate VSR performance, formulated as follows:

\vspace{-2mm}
\begin{equation}
\label{eq:cer}
    CER = \frac{S + D + I}{N}
\end{equation}
\vspace{-2mm}

\noindent where $S$, $D$, and $I$ represent the number of substitutions, deletions, and insertions in the output transcription, respectively. 
$N$ is the number of characters in the ground truth transcription.

\subsection{Data profile}
\vspace{-1mm}

CNVSRC 2023 utilized the CN-CVS dataset~\cite{chen_cn-cvs_2023} in addition to two supplementary datasets: CNVSRC-Single and CNVSRC-Multi, which served as the development and evaluation data for the single-speaker VSR task (T1) and multi-speaker VSR task (T2), respectively. 
All the data was transcribed into text. Table~\ref{tab:data} presents the data profile of the three datasets.
Note that both CNVSRC-Single and CNVSRC-Multi were split into a development set and an evaluation set. 
The development data was transparent to the participants (including the video, audio, and text), while 
the text and audio of the evaluation data were kept secret during the entire challenge process. 
The participants could use the development data in any way, 
e.g., freely splitting it into a subset for model fine-tuning and a subset for model validation/selection.

\begin{table}[!htb]
\centering
\caption{Data profile used in CNVSRC 2023.}
\vspace{-2mm}
\label{tab:data}
\resizebox{0.9\columnwidth}{!}{
    \begin{tabular}{l|c|cc|cc}
    \toprule
                    & CN-CVS  & \multicolumn{2}{c|}{CNVSRC-Single}  & \multicolumn{2}{c}{CNVSRC-Multi}  \\
        \midrule 
        DataSet     & Train   & Dev          & Eval                 & Dev       & Eval       \\
        \midrule 
        \# Spks     & 2,557   & \multicolumn{2}{c|}{1}              & \multicolumn{2}{c}{43}       \\      
        \midrule                                                                                                          
        \# Videos   & 206,261 & 25,947       & 2,881                & 20,450    & 10,269           \\ 
        \# Hours    & 308.00  & 94.00        & 8.41                 & 29.24     & 14.49            \\ 
    \bottomrule
    \end{tabular}}
    \vspace{-5mm}
\end{table}

\vspace{-1mm}
\subsubsection{CN-CVS}

The CN-CVS dataset~\cite{chen_cn-cvs_2023} comprises visual-speech data from over 2,557 speakers, totalling more than 300 hours of videos.
It encompasses various scenarios, including news broadcasts and public speeches.
So far, CN-CVS is the largest open-source Chinese visual-speech dataset. 
This dataset was used as the primary training data for the CNVSRC 2023 challenge.
Note that the original publication of CN-CVS is for the video-to-speech synthesis (VTS) task and thus does not involve text transcription. 
To support the CNVSRC 2023 challenge, we labelled the video with a semi-automatic pipeline that involves ASR transcribing and human check, 
as will be presented shortly. 

\vspace{-1mm}
\subsubsection{CNVSRC-Single}
\vspace{-1mm}

The CNVSRC-Single dataset, designed for a single-speaker VSR task (T1), is obtained from a broadcaster's online channel. 
It includes over 800 speech videos of that broadcaster, with a cumulative duration of over 100 hours. 
The pipeline used in~\cite{chen_cn-cvs_2023} was employed for the collection and processing.

\vspace{-1mm}
\subsubsection{CNVSRC-Multi}
\vspace{-1mm}

The CNVSRC-Multi dataset was designed as the development/evaluation data for the multi-speaker VSR task (T2). 
It encompasses two scenarios: reading in a recording studio and speeches downloaded from the internet. 

In the recording studio scenario, facial videos of speakers were captured from three different 
camera angles (0°,30°,60°), while their speech audio was recorded using a high-quality microphone. 
The speakers were prompted a sentence at each time via a computer screen and were asked to read the sentence clearly with a neutral emotion. 
The video data from the \emph{front camera} was used in CNVSRC 2023, 
which was transcoded to 25 frames per second (FPS) and scaled to an appropriate size. 
A total of 23 speakers participated in the recording, each reading 1,000 sentences. 
In the speeches from the internet scenario, videos of public speeches from 20 speakers 
were collected from the internet, again following the same collection and processing pipeline as~\cite{chen_cn-cvs_2023}.
To ensure the integrity of the collected videos from the internet, a face recognition tool\footnote{https://pypi.org/project/face-recognition/} was employed to check whether there is only one face in each video frame 
and if the face is the target face. 
A manual check was then conducted to double-check that each extracted video only contained the target face.

\subsection{Text annotation}
\vspace{-1mm}

To generate text transcriptions, a Paraformer-based ASR system~\cite{gao2022paraformer} was employed to transcribe the speech of all the videos. 
Furthermore, the manual check was conducted to ensure that the CER of the transcriptions remains below 2\%.

\vspace{-1mm}
\section{Baseline System}
\label{sec:baseline}

\begin{table*}[htb]
\centering
\vspace{-1.5mm}
\caption{Training Details of the Pretraining (P1 \& P2) and Fine-tuning (FT) steps when constructing the baseline systems.}
\vspace{-1.5mm}
\label{tab:training}
\resizebox{1.55\columnwidth}{!}{
\begin{tabular}{l|c|c|c|c}
    \toprule
    Experiment        & P1                     & P2              & FT (Single-Speaker)              & FT (Multi-Speaker)                  \\
    \midrule                                                                                                     
    Initialize        & Random                 & P1 Saved Model  & P2 Saved Model  & P2 Saved Model      \\
    Warmup Epochs     & 5 & 5 & 2 & 2\\
    Learning Rate     & 0.0002                 & 0.001           & 0.0003          & 0.0002              \\
    Training Epochs   & 75 + Early stop                     & 75              & 80              & 80                  \\
    Saved Model       & Top 10 average     & Last 10 epochs average & Last 5 epochs average  & Last 5 epochs  average     \\
    \bottomrule
\end{tabular}}
\vspace{-2mm}
\end{table*}

Leveraging the state-of-the-art model used in Auto-AVSR~\cite{ma_auto-avsr_2023}, we trained two baseline systems, 
one for the single-speaker VSR task (T1) and the other for the multi-speaker VSR task (T2). 
Only the datasets provided in this challenge were used. In other words, these systems conform to the specifications of the fixed tracks.

\subsection{Model structure}
\vspace{-1mm}

The model structure is duplicated from Auto-AVSR~\cite{ma_auto-avsr_2023}.
Specifically, it comprises three components: visual frontend, encoder, and decoder. 
As in~\cite{ma_auto-avsr_2023}, the visual frontend adopts ResNet18 as its backbone, except that the first 2D-CNN layer is replaced with a 3D-CNN layer to capture local spatiotemporal correlation. 
The encoder adopts a Conformer structure with 12 layers, while the decoder employs a Transformer structure with 6 layers.

Upon receiving the input video data, the visual frontend performs initial local spatiotemporal feature extraction. 
Subsequently, the Conformer encoder further extracts context-dependent features. 
A projection layer and a transformer decoder are employed to predict the output class labels. 
The entire model was trained with the joint CTC/Attention loss, where the CTC loss is back-propagated through the projection layer, 
while the Attention loss is back-propagated through the decoder. 
Refer to~\cite{ma_auto-avsr_2023} for details.

\subsection{Data pre-processing}
\vspace{-1mm}

The provided datasets of the challenge include the faces of the target speakers, 
so a pre-processing pipeline was designed to extract the lip region. 
Note that the pipeline was equally applied to both the training data (CN-CVS) 
and the development/evaluation data (CNVSRC-Single and CNVSRC-Multi).

Initially, we utilized RetinaFace~\cite{deng_retinaface_2019} to detect the facial regions in each frame. 
Subsequently, FAN~\cite{bulat_how_2017} was employed to extract facial landmarks, 
with which each detected face was aligned with a mean reference face. 
Finally, we extracted the lip region through the alignment for each video frame, which served as the input to the visual frontend.
The modelling units are subword tokens, and the tokenizer was trained using the SentencePiece~\cite{kudo_sentencepiece_2018} tool in the unigram mode. During the training of this tokenizer, only the text data from CN-CVS was utilized. 
Subsequently, we employed this tokenizer to process the CN-CVS, CNVSRC-Single, and CNVSRC-Multi datasets, 
obtaining the token sequences used to train the recognition models.

\subsection{Training strategy}
\vspace{-1mm}

We followed a two-step process to build the baseline systems. 
Firstly, we performed pre-training with the CN-CVS dataset. 
Subsequently, the development set was split into an `adaptation set' and a `validation set', 
with a ratio of 8:1 for the single-speaker dataset and 3:1 for the multi-speaker dataset. 
The adaptation set was used to fine-tune the pre-trained model, while the validation set was used to select the appropriate checkpoint. 
The training process was summarized in Table~\ref{tab:training}, and the details are as follows. 

The initial training phase (P1) selected CN-CVS videos with a duration of less than 4 seconds to train an `easy model'. 
The maximum number of training epochs was 75, and early stopping was triggered if the model's performance on the validation set started to drop. Once the training stopped, we selected the 
top 10 models based on their accuracy on the validation set, averaged their parameters to obtain the P1 model. 
Next, a full pre-training phase (P2) was evoked using the complete CN-CVS dataset, and the training was conducted for 75 epochs. 
Note that a warmup stage of 5 epochs was designed, by which the learning rate was gradually increased from 0 to 0.001. 
The average of the models of the last 10 epochs was used as the P2 model, i.e., the pre-trained model.

The fine-tuning step started from the P2 model and ran 80 epochs, including 2 epochs of warmup. 
This process is the same for the models trained for the single-speaker task and the multi-speaker task, 
but there are indeed some differences. 
Besides the learning rate (see Table~\ref{tab:training}), the most notable difference is that for the single-speaker model, 
we randomized the parameters of the classification layer of the P2 model, 
to provide sufficient space for the adaptation with the large amount of single-speaker data. 
The average of the models of the last 5 epochs was used as the final model, 
for both the single-speaker task and the multi-speaker task.



\subsection{Performance}
\vspace{-1mm}

The performance of the baseline models was evaluated on the respective validation set and evaluation set for both the single-speaker task and multi-speaker task, using the TorchMetrics tool\footnote{https://lightning.ai/docs/torchmetrics/stable/}. 
The CER results are as shown in Table~\ref{tab:task}.

\begin{table}[htb]
\centering
\footnotesize
\caption{Performance of the baseline systems.}
\vspace{-1.5mm}
\label{tab:task}
\resizebox{0.9\columnwidth}{!}{
\begin{tabular}{c|c|c}
    \toprule
                       & \multicolumn{2}{c}{Character Error Rate}  \\
    \midrule 
    Task               & T1: Single-speaker VSR   &  T2: Multi-speaker VSR   \\
    \midrule 
    Valid              & 48.57\%   & 58.77\%    \\
    Eval               & 48.60\%   & 58.37\%    \\
    \bottomrule
\end{tabular}}
\vspace{-2mm}
\end{table}

\begin{table*}[htb]
\centering
\footnotesize
\caption{Leaderboard of CNVSRC2023. Team ID and CER are reported.}
\vspace{-1.5mm}
\label{tab:result}
\begin{tabular}{l|cc|cc|cc|cc}
    \toprule
    Task       & \multicolumn{4}{c|}{T1: Single-speaker VSR}                 & \multicolumn{4}{c}{T2: Multi-speaker VSR}    \\
    \midrule
    Track      & \multicolumn{2}{c|}{Fixed Track}&\multicolumn{2}{c|}{Open Track}&\multicolumn{2}{c|}{Fixed Track}&\multicolumn{2}{c}{Open Track} \\
    \midrule
    Baseline   &\multicolumn{2}{c|}{48.60\%}  &\multicolumn{2}{c|}{48.60\%}  &\multicolumn{2}{c|}{58.37\%}  &\multicolumn{2}{c}{58.37\%}           \\
    \midrule
    Rank 1     & T237 & 34.76\%             & T237 & 34.76\%             & T244 & 53.68\%             & T237 & 41.05\%           \\
    Rank 2     & T266 & 38.09\%             &      &                       & T267 & 54.56\%           & T244 & 53.68\%         \\
    Rank 3     & T290 & 39.47\%             &      &                       &      &                   &      &                 \\
    Rank 4     & T238 & 40.52\%             &      &                       &      &                   &      &                 \\
    Rank 5     & T267 & 41.62\%             &      &                       &      &                   &      &                 \\
    \bottomrule
\end{tabular}
\vspace{-3mm}
\end{table*}

\vspace{-1mm}
\section{CNVSRC 2023 Report}
\label{sec:result}

\subsection{Leaderboard}
\vspace{-1mm}

CNVSRC 2023 received 10 valid submissions from 6 teams. 
Most teams chose to submit their results to the single-speaker task, 
suggesting that single-speaker VSR is more suitable for the current stage of technical development, and multi-speaker VSR is over-challenging. The leaderboard is reported in Table~\ref{tab:result}.

Overall, T237 achieved the best performance in 3/4 of the tasks \& tracks, 
and their results outperformed the baseline systems by a large margin.
Their proposed system consists of a ResNet-3D visual frontend, an E-Branchformer encoder~\cite{kim2023branchformer}, and a Transformer decoder.
The Chinese characters were used as the modelling units, and multiple data augmentation methods, including speed perturbation, random rotation, and horizontal flipping, were applied during training.
Additionally, a ROVER-based system fusion~\cite{fiscus1997post} was performed during the inference procedure.

\vspace{-1mm}
\subsection{Technical summary}
\vspace{-1mm}
\label{sec:technical_summary}

We summarize here the promising techniques demonstrated by the results of the submissions, highlighting the most effective methods in \textbf{bold font}.

\vspace{-1.5mm}
\subsubsection{Data pre-processing}
\vspace{-1mm}

Many teams adhered to the baseline system for data pre-processing. A notable observation is that T237 extracted lip regions of varying sizes and resolutions as inputs to their model and discovered that \textbf{larger regions (lip + mouth around)} yielded clear performance improvement.

\vspace{-1.5mm}
\subsubsection{Data augmentation}
\vspace{-1mm}

The participants extensively used data augmentation techniques, including random erase, random crop, random flip, and adaptive time masking. 
Notably, \textbf{speed perturbation} and \textbf{generative data augmentation} were reported to yield unexceptionally remarkable results.
Speed perturbation adjusts the speed of the original videos by a factor ranging from 0.9 to 1.1. 
T237 and T238 observed notable enhancements in model performance (4.86\% relative improvement) by applying speed perturbation.
Generative data augmentation involves generating speech-driven lip videos, hence producing extra video-text training pairs. 
T238 utilized facial images from CNVSRC-Single and speech from CN-CVS and CNVSRC-Single to create an extra set of video-text pairs and reported a relative CER reduction of 6.98\%.

\vspace{-1.5mm}
\subsubsection{Model structure}
\vspace{-1mm}

Most of the participating teams adopted the model architecture of the baseline systems, though some teams chose a more complicated backbone to pursue better performance.
For instance, T237 achieved superior results using a \textbf{ResNet3D} structure. Moreover, T237 employed two advanced encoder structures: Branchformer~\cite{peng2022branchformer} and \textbf{E-Branchformer}~\cite{kim2023branchformer}. All these advanced structures lead to notable performance improvements.
T266 introduced an \textbf{inner CTC residual module}~\cite{nozaki2021relaxing,lee2021intermediate,burchi2023audio} that resides in the Conformer block of the encoder.
This module back-propagated a CTC loss through the shallow layers thus facilitating more effective parameter updates for the shallow layers of the model.
Furthermore, taking inspiration from~\cite{wu2021u2++, zhang2022wenet}, T266 utilized a \textbf{bi-transformer} structure to construct the decoder.
This modification enhances the model's ability to capture contextual information from both the past and future segments.

\vspace{-1.5mm}
\subsubsection{Modeling units}
\vspace{-1mm}

Most participating teams used \textbf{Chinese characters as the modelling units}, and showed better performance than the subword tokens used by the baseline systems. 
In addition to subword tokens, T238 used \textbf{phonemes as supplementary modelling units} and designed a separate decoder for phoneme recognition. This approach achieved performance improvement, for which a hypothesis is that phonemes contain less semantic variation and thus are more closely related to lip movement, which may stabilize the training, especially in the early stage. 

\vspace{-1.5mm}
\subsubsection{Cross-modality modeling}
\vspace{-1mm}

Some teams designed various approaches to \textbf{leverage the cross-modal dependency}.
T290 invented an ASR-VSR joint system that forces the video representations to predict not only the text labels but also the speech representations produced by the middle layer of the ASR system.
Following the same inspiration, T244 trained an audio-visual recognition system where the ASR and VSR have their respective independent encoders and decoders, and an AVSR decoder is constructed on top of the fused ASR and VSR features. 

%

\vspace{-1.5mm}
\subsubsection{Decoding strategy}
\vspace{-1mm}

Several teams integrated RNN-based or Transformer-based language models to enhance the decoder, and mild performance improvement was reported. 
Moreover, system fusion was widely employed by teams to improve the performance of their systems.

\vspace{-1mm}
\section{Conclusion}
\label{sec:conclusion}

This paper comprehensively details the inaugural Chinese Continuous Visual Speech Recognition Challenge (CNVSRC 2023).
A key motivation of the challenge is to investigate the performance bound of VSR under the present data resource, e.g., 300 hours of training data from 2,557 speakers. The overall results suggest that the performance is far from satisfactory, even for the single-speaker scenario where about 100 hours of video is available for one person. The poor performance is certainly attributed to the lack of data, but whether it is related to the special linguistic properties of Chinese, e.g., the ubiquitous homophones, is unknown. 

Based on these technical reports from participants, we have summarized the key techniques that might be crucial in constructing Chinese VSR systems. The most effective methods, ordered by their merit in terms of CER reduction: \textbf{Chinese characters as modelling units}, \textbf{rich data augmentation}, \textbf{fully 3D-CNN visual frontend}, \textbf{cross-modality modelling}, \textbf{system fusion}.
Leveraging the technical insights provided by the participants, we have established a cutting-edge benchmark for Chinese LVC-VSR. We aspire that these resources will strengthen the burgeoning field of LVC-VSR research.

\newpage
\bibliographystyle{IEEEtran}
\bibliography{main}

\end{document}